\documentclass{article}

    \PassOptionsToPackage{numbers}{natbib}

\usepackage[preprint]{neurips_2020}

\usepackage{natbib}
\usepackage[utf8]{inputenc} 
\usepackage[T1]{fontenc}    
\usepackage{hyperref}       
\usepackage{url}            
\usepackage{booktabs}       
\usepackage{amsfonts}       
\usepackage{nicefrac}       
\usepackage{microtype}      
\usepackage{comment}

\usepackage{hyperref}
\usepackage{geometry}
\usepackage{graphicx}
\usepackage{upgreek}
\usepackage{url}
\usepackage{cite}
\usepackage{longtable}
\usepackage{cellspace, makecell, multirow}
\usepackage{subcaption}
\usepackage{amsmath,amsfonts,amssymb,bm}
\usepackage{amsthm}
\usepackage{multirow}
\usepackage{booktabs, multirow, array, makecell, caption}
\usepackage{algpseudocode}
\usepackage[vlined, ruled, boxed, linesnumbered]{algorithm2e}
\usepackage{indentfirst} 
\usepackage{breqn}
\newtheorem{theorem}{Theorem}

\usepackage{soul}
\let\existstemp\exists
\let\foralltemp\forall
\renewcommand*{\exists}{\existstemp\mkern2mu}
\renewcommand*{\forall}{\foralltemp\mkern2mu}

\DeclareMathOperator*{\argmin}{arg\,min}

\theoremstyle{definition}

\usepackage{xcolor}

\newcommand{\veryshortarrow}[1][3pt]{\mathrel{%
   \vcenter{\hbox{\rule[-.5\fontdimen8\textfont3]{#1}{\fontdimen8\textfont3}}}%
   \mkern-4mu\hbox{\usefont{U}{lasy}{m}{n}\symbol{41}}}}

\makeatletter

\setbox0\hbox{$\xdef\scriptratio{\strip@pt\dimexpr
    \numexpr(\sf@size*65536)/\f@size sp}$}

\newcommand{\scriptveryshortarrow}[1][3pt]{\mathrel{%
    \vcenter{\hbox{\rule[-.5\fontdimen8\scriptfont3]
               {\scriptratio\dimexpr#1\relax}{\fontdimen8\scriptfont3}}}%
   \mkern-4mu\hbox{\let\f@size\sf@size\usefont{U}{lasy}{m}{n}\symbol{41}}}}

\makeatother

\usepackage{appendix}

\title{Preserving Domain Private Representation via Mutual Information Maximization}

\author{%
  Jiahong Chen\thanks{Both authors contributed equally to this work.} \\
  Department of Mechanical Engineering\\
  The University of British Columbia\\
  Vancouver, Canada\\
  \texttt{jhchen@mech.ubc.ca} \\
   \And
   Jing Wang\footnotemark[1] \\
   Department of Mechanical Engineering \\
   The University of British Columbia\\
  Vancouver, Canada\\
   \texttt{jing@ece.ubc.ca} \\
   \AND
   Weipeng Lin \\
   School of Artificial Intelligence\\
   Shenzhen Polytechnic\\
   Shenzhen, Guangdong, China\\
   \texttt{weipengl@szpt.edu.cn} \\
   \And
   Kuangen Zhang \\
   Department of Mechanical Engineering \\
   The University of British Columbia\\
  Vancouver, Canada\\
   \texttt{kuangen.zhang@alumni.ubc.ca} \\
   \And
   Clarence W. de Silva \\
   Department of Mechanical Engineering\\
  The University of British Columbia\\
  Vancouver, Canada\\
   \texttt{desilva@mech.ubc.ca} \\
}

\begin{document}

\maketitle

\begin{abstract}

Recent advances in unsupervised domain adaptation have shown that mitigating the domain divergence by extracting the domain-invariant representation could significantly improve the generalization of a model to an unlabeled data domain. Nevertheless, the existing methods fail to effectively preserve the representation that is private to the label-missing domain, which could adversely affect the generalization. In this paper, we propose an approach to preserve such representation so that the latent distribution of the unlabeled domain could represent both the domain-invariant features and the individual characteristics that are private to the unlabeled domain. In particular, we demonstrate that maximizing the mutual information between the unlabeled domain and its latent space while mitigating the domain divergence can achieve such preservation. We also theoretically and empirically validate that preserving the representation that is private to the unlabeled domain is important and of necessity for the cross-domain generalization. Our approach outperforms state-of-the-art methods on several public datasets.

\end{abstract}

\section{Introduction}
Supervised learning algorithms highly rely on a considerable amount of labeled data, as more labeled data will introduce a more discriminative feature representation and lead to a better performance. However, in real-case applications, labeling data for every specific scenario is costly. A common approach to solve this problem is to assume that a related public dataset covers sufficient variations of the data distribution, and to attempt to generalize a discriminative model directly from it. However, the data distribution of one dataset is typically different from that of another dataset. Therefore, there is a strong demand to design some feature extraction strategies that can ultimately migrate the domain shift between the labeled dataset (source domain) and the label-missing dataset (target domain). In response to this, unsupervised domain adaptation (UDA) methods are proposed to mitigate the domain shift by either extracting the features that are invariant to such domain shift or by aligning the feature distributions from one domain to the other. 

In this work, we focus on the tasks of object classification, which have numerous benchmark datasets that allow us to better compare our proposed work with the existing methods and validate the importance of this research. Previous work mainly focuses on exploring different ways to mitigate the domain shift between the two data domains. While mitigating the domain shift can learn a shared representation that is important to a specific task, performing adaptation in this manner alone ignores the specific properties of the target domain, which is also important to the task on the target domain. To weaken the side effect of the shared-representation learning to the specific task on the target domain, the representation that is private to the target domain is required to be preserved while trying to mitigate the domain shift. 

To achieve this goal, we propose to preserve the representation that is private to the target domain through maximizing the non-linear statistical dependencies, i.e., mutual information (MI), between the target sampling space and its representation (latent) space. Recall that the objective of UDA is to infer a reasonable latent distribution for the target domain, which can utilize the discriminative features learned from the source domain for its own task. Therefore, this distribution alignment, which is particularly for the UDA problem, can be achieved by minimizing the difference between the discriminative source latent distribution and the conditional distribution of the target latent space given its input space. One of the choices for quantifying such difference would be the KL-divergence. The mutual information maximization is, in fact, derived from this KL-divergence penalty, which theoretically validates the importance of preserving the representation that is private to the target domain in solving the UDA problem. Unfortunately, the conditional distribution of the target latent space given its input space is intractable as the correspondence between the target latent samples that can be used for accurately predict their labels and the target input samples is unknown. To tackle this issue, we can derive a lower bound so that maximizing the lower bound will be equivalent to minimizing the KL-divergence penalty. 

Another problem is that the mutual information between high dimensional continuous random variables is difficult to estimate \citep{paninski2003estimation}. The existing non-parametric methods to exactly compute mutual information is only tractable for a limited family of problem where the probability distributions are known \citep{fraser1986independent, hulle2005edgeworth, kwak2002input, suzuki2008approximating}. However, our goal is to utilize the maximization of mutual information to preserve the information about one known probability distribution, i.e., the target input distribution, in another unknown probability distribution, i.e., the target latent distribution. We should rely on a more general approach for the mutual-information estimation. Fortunately, a general-purpose estimator for mutual information, namely Mutual Information Neural Estimator \citep{belghazi2018mutual}, is proposed recently, which utilizes the Donsker-Varadhan representation of the KL-divergence between the joint distribution and the product of the marginals to estimate the MI. 

Here we introduce a novel method, \emph{Domain Preservation Nets} (DPN), for preserving the representation that is private to the target domain. Unlike the existing work to solve the UDA problem, we explicitly unveil that it is important to preserve the representation that is private to the target domain. We validate this claim both theoretically and empirically. The objective to preserve the private representation for the target domain is, in fact, a part of the objective that minimizes the divergence between the source latent distribution and the intractable target latent distribution to achieve the knowledge transfer. We further transform the intractable part into a lower bound that can be maximized through optimization. Experimental results on several benchmark datasets suggest that DPN outperforms state-of-the-art methods on several UDA tasks. The experiments on the third dataset also show the superior generalization of our model to an unknown scenario.

\section{Related Work}
\subsection{Domain Adaptation}

In this literature review, we focus on Convolutional Neural Network (CNN) based methods due to their empirical superiority on the UDA problem. Ben-David et al. \citep{ben2010theory} unveil that the success of learning a decision rule that can be applied to the label-missing domain is theoretically achieved by deriving a convergence learning bound in terms of the classification error on the source domain and the decision divergence on the classification, i.e., the classifier-induced discrepancy, in the target domain. Most of the existing UDA methods that are developed based on the domain-invariant feature extraction follow this theoretical insight, and attempt to measure such decision divergence using the features that are induced by classifiers \citep{borgwardt2006integrating, gretton2012kernel, tzeng2015simultaneous, saito2018maximum}.

Adversarial domain adaptation methods are motivated by generative adversarial networks (GANs) \citep{goodfellow2014generative}. Domain-adversarial neural network (DANN) proposes a domain-adapted classifier to quantify the classifier-induced discrepancy \citep{ganin2016domain}. The adversarial learning strategies make the training of such classifier more difficult to converge to promote the domain-invariant feature extraction \citep{sankaranarayanan2018generate, pei2018multi,cao2018partial,long2018conditional, liu2019transferable, saito2018maximum, xie2018learning}. Non-adversarial domain adaptation is another important direction of UDA. The existing methods that are developed in this direction attempt to quantify the domain shift by certain statistical distances between the source domain and the target domain. Methods based on maximum mean discrepancy (MMD), which measure the variance between the feature distributions that are extracted from the two domains, are proposed \citep{borgwardt2006integrating, tolstikhin2016minimax, long2015learning, long2017deep}. The difference of the mean and the covariance between the input spaces of the two domains can also be used to measure the domain shift \citep{sun2016return, sun2016deep}. Recent studies also unveil that the domain shift relies on the less-informative features, which has small norms for the target-specific task \citep{xu2019larger}. 


\subsection{Mutual Information}
The work of Denoising AutoEncoder (DAE) \citep{vincent2008extracting, vincent2010stacked} shows that an autoencoder that is regularized to minimize the reconstruction error can maximize the lower bound of the mutual information between the input space and its latent representation. Deep Reconstruction-Classification Network (DRCN) \citep{ghifary2016deep} then utilizes this principle to make the encoding process focus on the information that benefits the knowledge transfer between the two domains. Domain separation network (DSN) \citep{bousmalis2016domain} further enforces this principle by preserving the representations that are private to the source domain and the target domain before the alignment.

Although an autoencoder can be used to maximize a lower bound of the mutual information between the input space and its latent representation, approximating the mutual information by reconstruction is not that effective and efficient. Especially when the principal task is not to reconstruct the input samples, the decoding process can unnecessarily consume a significant amount of computing resources. Instead of approximating a task-specific mutual information, Mutual Information Neural Estimator (MINE) \citep{belghazi2018mutual} is proposed as a general-purpose estimator to approximate the mutual information by estimating the tight lower bound of the mutual information \citep{ruderman2012tighter} by taking advantage of the Donsker-Varadhan representation \citep{donsker1983asymptotic}. To estimate the lower bound, MINE utilizes a shared-weight deep neural network to jointly estimate the joint probability of the two random variables and the product of their marginals. 

\section{Methodology}
\label{sec:method}
In the setting of UDA, the labels of the target input samples are not accessible. The objective of an UDA method is to infer a latent distribution from the target domain with the support of the discriminative source features so that the latent distribution can be used to accurately predict those unlabeled input samples. In this paper, we denote the source domain as $D_S=\{(\mathbf{x}^{(i)}_s, \mathbf{y}^{(i)}_s)\}_{i=1}^n$, which consists $n$ pairs of image and label that are sampled from the source input space $\{X_S, Y_S\}$. Similarly, the target domain, which is denoted as $D_T=\{(\mathbf{x}^{(j)}_t)\}_{j=1}^m$, provides $m$ unlabeled images sampled from the target input space $X_T$.




\subsection{Unsupervised Domain Adaptation as Distribution Alignment}\label{sec:distribution-alignment}

The most direct way to transfer the knowledge across two data domains is to align the latent distributions that are inferred from the two domains. The distribution alignment can then be achieved by minimizing the Kullback-Leibler divergence between the two distributions \citep{kingma2013auto},
\begin{equation}
    D_{KL}(\mathbb{Q}||\mathbb{P}) = \mathbb{E}_{\mathbb{Q}}[\log \frac{d\mathbb{Q}}{d\mathbb{P}}],
\end{equation}
where the distributions $\mathbb{P}$ and $\mathbb{Q}$ have to be defined on the same probability space to make the KL-divergence penalty finite.

In the single-domain learning, the probability space of the latent representation is explored by optimizing the conditional probability distribution of a latent variable $\mathbf{z}$ given its observation $\mathbf{x}$ \citep{blei2017variational}. A way to achieve this is to approximate the conditional probability distribution $\mathbb{P}(\mathbf{z} | \mathbf{x})$ by a tractable probability distribution $\mathbb{Q}(\mathbf{z})$,
\begin{equation}\label{eqn:kl-divergence}
    \mathbb{Q}^{*}(\mathbf{z}) = \argmin D_{KL}(\mathbb{Q}(\mathbf{z})||\mathbb{P}(\mathbf{z} | \mathbf{x})).
\end{equation}


In the case of the cross-domain learning, the discriminative latent distribution of the source domain, $\mathbb{Q}(\mathbf{z}_s)$, is assumed to be tractable because the labels of the source input samples are accessible. On the contrary, the target input samples are not labeled for the specific task, and thus the discriminative latent distribution of the target domain is intractable. In order to infer a discriminative latent distribution from the target domain with the support of the source discriminative features, the target latent distribution should be approximated to be closed to the discriminative source latent distribution,
\begin{equation}\label{eqn:kld-obj}
    \mathbb{Q}^{*}(\mathbf{z}_t) = \argmin D_{KL}(\mathbb{Q}(\mathbf{z}_s)||\mathbb{P}(\mathbf{z}_t | \mathbf{x}_t)).
\end{equation}

Equation \ref{eqn:kld-obj} can be regarded as the objective function to solve the UDA problem. Note that $\mathbb{P}(\mathbf{z}_t | \mathbf{x}_t)$ is intractable because the target labels are unknown; thus the correspondence between the target latent vectors for classification and the target input samples is also unknown. To solve this issue, Theorem \ref{theo:evidence-loewr-bound} is derived to optimize the Equation \ref{eqn:kld-obj} by finding its evidence lower bound.

\begin{theorem}\label{theo:evidence-loewr-bound}
The evidence lower bound for minimizing $D_{KL}(\mathbb{Q}(\mathbf{z}_s)||\mathbb{P}(\mathbf{z}_t | \mathbf{x}_t))$ is
\begin{equation}\label{eqn:theo-elbo}
        \mathcal{B}(\mathbb{P}) = \mathbb{E}[\log \mathbb{P}(\mathbf{x}_t|\mathbf{z}_t)] - D_{KL}(\mathbb{Q}(\mathbf{z}_s)||\mathbb{P}(\mathbf{z}_t));
\end{equation}
therefore maximizing $\mathcal{B}(\mathbb{P})$ is equivalent to minimizing $D_{KL}(\mathbb{Q}(\mathbf{z}_s)||\mathbb{P}(\mathbf{z}_t | \mathbf{x}_t))$,
\begin{equation}\label{eqn:theo-elbo-equivalent}
    \begin{split}
        \min D_{KL}(\mathbb{Q}(\mathbf{z}_s)||\mathbb{P}(\mathbf{z}_t | \mathbf{x})) 
        =& \max \mathcal{B}(\mathbb{P}).
    \end{split}
\end{equation}
\end{theorem}



The proof of Theorem \ref{theo:evidence-loewr-bound} is in supplementary material. Combining Equation \ref{eqn:theo-elbo} and Equation \ref{eqn:theo-elbo-equivalent}, minimizing $D_{KL}(\mathbb{Q}(\mathbf{z}_s)||\mathbb{P}(\mathbf{z}_t | \mathbf{x}_t))$ requires to minimize $D_{KL}(\mathbb{Q}(\mathbf{z}_s)||\mathbb{P}(\mathbf{z}_t))$, which encourages the approximation of the target latent distribution to be close to the discriminative source latent distribution. Apart from this, the expectation of the conditional probability $\mathbb{E}[\log \mathbb{P}(\mathbf{x}_t|\mathbf{z}_t)]$ is required to be maximized, which preserves the representation that is private to the target domain. Theorem \ref{theo:infomax-equivalent} addresses the maximization of $\mathbb{E}[\log \mathbb{P}(\mathbf{x}_t|\mathbf{z}_t)]$ in terms of mutual information. 

\begin{theorem}\label{theo:infomax-equivalent}
Maximizing the expectation of the conditional probability $\mathbb{E}[\log \mathbb{P}(\mathbf{x}_t|\mathbf{z}_t)]$ is equivalent to maximizing the mutual information between $X_T$ and $Z_T$,
\begin{equation}\label{eqn:theo-infomax-equivalent}
    \max \mathbb{E}[\log \mathbb{P}(\mathbf{x}_t|\mathbf{z}_t)] = \max \mathsf{I}(X_T;Z_T).
\end{equation}
\end{theorem}




The proof of Theorem \ref{theo:infomax-equivalent} is in supplementary material. Combining Equation \ref{eqn:theo-elbo}, Equation \ref{eqn:theo-elbo-equivalent}, and Equation \ref{eqn:theo-infomax-equivalent}, we have
\begin{equation}\label{eqn:evidence-lower-bound-kl-mi}
    \max \mathcal{B}(\mathbb{P}) = \max \mathsf{I}(X_T;Z_T) - D_{KL}(\mathbb{Q}_{Z_S}(\mathbf{z}_s)||\mathbb{P}_{Z_T}(\mathbf{z}_t)),
\end{equation}
where the maximization of the mutual information preserves the representation that is private to the target domain; and the minimization of the KL-divergence transfers the knowledge from the source domain to the target domain.
Therefore, by maximizing the lower bound shown in Equation \ref{eqn:evidence-lower-bound-kl-mi}, the domain private representation can be preserved while transferring the knowledge.

\subsection{Mutual Information Estimation for Private Representation Preservation}
As presented in Section \ref{sec:distribution-alignment}, maximizing the mutual information can maximize the likelihood of the input samples conditioned on their latent representation, which can be used to preserve the domain private representation. Mutual information can capture the non-linear statistical dependencies between two random variables by quantifying the level of difference between the joint distribution of the two random variables and the product of their marginal distributions. Denote the MI between two random variables $X$ and $Z$ as
\begin{equation}\label{eqn:infomax}
    \mathsf{I}(X;Z) = \mathsf{H}(X) -  \mathsf{H}(X|Z),
\end{equation}
where $\mathsf{I}(\cdot)$ denotes the MI between the sampling spaces of the two random variables, and $\mathsf{H}(\cdot)$ denotes the Shannon entropy. Let $(X,Z)$ be a pair of random variables with their values over the space $\mathcal{X}\times \mathcal{Z}$. We denote their joint distribution as $\mathbb{P}_{XZ}$ and their marginal distributions as $\mathbb{P}_X$ and $\mathbb{P}_{Z}$. The MI can be defined as the KL-divergence between $\mathbb{P}_{XZ}$ and the product of the marginals $\mathbb{P}_X \otimes \mathbb{P}_Z$:
\begin{equation}\label{eqn:mi-to-kld}
    \mathsf{I}(X;Z) = D_{KL}(\mathbb{P}_{XZ} || \mathbb{P}_X \otimes \mathbb{P}_Z).
\end{equation}

Based on \citep{donsker1983asymptotic, belghazi2018mutual}, the KL-divergence admits the following Donsker-Varadhan representation:
\begin{equation}\label{eqn:kld-sup}
    D_{KL}(\mathbb{P}_{XZ}||\mathbb{P}_X \otimes \mathbb{P}_Z) = \sup_{M:\mathcal{X}\times\mathcal{Y}\rightarrow\mathbb{R}}\mathbb{E}_{\mathbb{\mathbb{P}_{XZ}}}[M]-\log (\mathbb{E}_{\mathbb{\mathbb{P}_X \otimes \mathbb{P}_Z}}[e^{M}]).
\end{equation}

Mutual Information Neural Estimator (MINE) is a deep neural network based method to estimate the lower bound of Equation \ref{eqn:kld-sup} by gradient descent \citep{belghazi2018mutual}. However, MINE utilizes a shared-weight artificial neural work to learn the joint probability and the product of the marginal probabilities jointly, which cannot model the joint and the product effectively for estimating the MI in the case of UDA. To tackle this problem, we propose to model the joint probability and the product of the marginal probabilities by two individual mapping functions that are trained separately. Let $\mathcal{H}$ contain any functions $M_1$ and $M_2$ such that $\forall M_1, M_2 \in \mathcal{H}, M_{1}:\mathcal{X}\times\mathcal{Y}\rightarrow\mathbb{R}$, $M_{2}:\mathcal{X}\times\mathcal{Y}\rightarrow\mathbb{R}$; then a lower bound for $D_{KL}(\mathbb{P}_{XZ}||\mathbb{P}_X \otimes \mathbb{P}_Z)$ can be derived based on the \textit{compression Lemma} in \citep{banerjee2006bayesian}.


\begin{theorem}\label{theo:two-fun-lower-bound}
For any two mapping functions $M_1, M_2 \in \mathcal{H}: \mathbb{E}_{\mathbb{P}_{XZ}}[M_2]\geq  \mathbb{E}_{\mathbb{P}_{XZ}}[M_1]$, the following bound yields:
\begin{equation}\label{eqn:two-fun-lower-bound}
    D_{KL}(\mathbb{P}_{XZ}||\mathbb{P}_X \otimes \mathbb{P}_Z)\geq \frac{1}{2}(\mathbb{E}_{\mathbb{P}_{XZ}}[M_1] - \log \mathbb{E}_{\mathbb{P}_X \otimes \mathbb{P}_Z}[e^{M_2}] ).
\end{equation}
\end{theorem}

The proof of Theorem \ref{theo:two-fun-lower-bound} is in supplementary material. Let $\mathcal{H}$ be defined on a class of functions of $M^{\theta}$, where $M^{\theta}:\mathcal{X}_T\times\mathcal{Z}_T\rightarrow\mathbb{R}$ is a deep neural network that is parameterized by $\theta$. Combining \autoref{eqn:mi-to-kld} and \autoref{theo:two-fun-lower-bound}, the lower bound for the mutual information can be derived as $\mathsf{I}(X_T;Z_T)\geq \mathsf{I}_{\Theta}(X_T;Z_T)$, which can be maximized by
\begin{equation}\label{eqn:mi-dnn-lower-bound}
\begin{split}
    \max_{\Theta} \mathsf{I}_{\Theta}(X_T;Z_T)
    \rightarrow &\max_{\theta_1, \theta_2}\frac{1}{2}(\mathbb{E}_{\mathbb{P}_{X_TZ_T}}[M_1^{\theta_1}]-\log \mathbb{E}_{\mathbb{P}_{X_T} \otimes \mathbb{P}_{Z_T}}[e^{M_2^{\theta_2}}]),
\end{split}
\end{equation}
where $\theta_1$ and $\theta_2$ denote the learning parameters that parameterize $M_1$ and $M_2$, respectively; the expectations are calculated using samples from $\mathbb{P}_{X_TZ_T}$ and $\mathbb{P}_{X_T} \otimes \mathbb{P}_{Z_T}$. The detailed implementation of Equation \ref{eqn:mi-dnn-lower-bound} is provided in Algorithm \ref{alg:da-mine}. 

\begin{algorithm}[t]
\SetAlgoLined
 Input image normalization; $(\theta_1, \theta_2)\gets$ network parameter initialization\;
 \While{epoch $\leq$ max epoch}{
  \For{$batch\gets1$ \KwTo N}{
   Draw $n$ minibatch samples $\mathbf{x}_t, \mathbf{z}_t$ from the joint distribution: \\
   ($\mathbf{x}_t^{(i)}$, $\mathbf{z}_t^{(i)}$) $\sim \mathbb{P}_{X_TZ_T}$, where $i \in [1,n]$\;
   
   Draw $n$ minibatch samples $\mathbf{\bar{z}}_t$ from the marginal distribution: \\
    $\mathbf{\bar{z}}_t^{(i)}$ $\sim \mathbb{P}_{Z_T}$, where $i \in [1,n]$\;
  
   Fix $G$ and $F$, and update $M_1$ and $M_2$ to $ \min_{M_1,M_2}[\mathcal{L}_{mi}(\mathbf{x}_t, \mathbf{z}_t, \mathbf{\bar{z}}_t)]$\;

   Calculate $\mathcal{L}_{cls}$, $\mathcal{L}_{kld}$, and $\mathcal{L}_{mi}$ using the current model parameters\;
   Fix $M_1$ and $M_2$, and update $G$ and $F$ to $\min_{G, F}[\mathcal{L}_{cls}+\alpha \mathcal{L}_{kld}+\beta \mathcal{L}_{mi}]$.
   }
 }
 \caption{DPN}
\label{alg:da-mine} 
\end{algorithm}

\subsection{Framework}
In this work, we follow the standard UDA protocol to use the softmax cross-entropy loss to evaluate the source-domain classification. To be specific, both the feature extractor $G$ and the image classifier $F$ are trained to minimize this objective function:
\begin{equation}
\begin{aligned}
\mathcal{L}_{cls}(X_{S}, Y_{S}) =  -\frac{1}{N} \sum_{i=1}^{N} \delta(\mathbf{y_s}^{(i)}) \log{[\sigma\circ F\circ G(\mathbf{x}_s^{(i)})]},
\end{aligned}
\end{equation}
where $G$ encodes the input samples to their latent representation ($\mathbf{z_s}^{(i)} = G(\mathbf{x_s}^{(i)})$); $\delta(\mathbf{y_s}^{(i)})$ is a binary indicator, which outputs 1 if the prediction $\mathbf{y_s}^{(i)}$ matches its corresponding class label $i$; $\sigma$ is the softmax function.

Then, combining Equation \ref{eqn:evidence-lower-bound-kl-mi} and Equation \ref{eqn:mi-dnn-lower-bound}, we have the lower bound $\mathcal{B}(\mathbb{P})$:
\begin{equation}
    \max \mathcal{B}(\mathbb{P}) = \max_{\theta_1, \theta_2}\frac{1}{2}(\mathbb{E}_{\mathbb{P}_{X_TZ_T}}[M_1^{\theta_1}]-\log (\mathbb{E}_{\mathbb{P}_{X_T} \otimes \mathbb{P}_{Z_T}}[e^{M_2^{\theta_2}}])) - D_{KL}(\mathbb{Q}_{Z_S}(\mathbf{z}_s)||\mathbb{P}_{Z_T}(\mathbf{z}_t)),
\end{equation}
where the KL-divergence penalty is utilized to regularize $G$ only. Therefore, the lower bound $\mathcal{B}(\mathbb{P})$ involes two terms: 1) a loss function $\mathcal{L}_{kld}$ to minimize the KL-divergence penalty between $Z_S$ and $Z_T$; 2) a mutual information penalty $\mathcal{L}_{mi}$, which optimizes $M_1$ and $M_2$, is used for preserving the representation that is private to the target domain. We formulate the objective function to optimize the mutual information as:
\begin{equation}
    \mathcal{L}_{mi}(\mathbf{x}_t, \mathbf{z}_t, \mathbf{\bar{z}}_t)=-\frac{1}{n}\sum_{i=1}^n [M_1^{\theta_1}(\mathbf{x}^{(i)}_t, \mathbf{z}^{(i)}_t)-\log (\frac{1}{n}\sum_{i=1}^n e^{M_2^{\theta^2}(\mathbf{x}^{(i)}_t, \mathbf{\bar{z}}^{(i)}_t)})],
\end{equation}
where $\mathbf{x}_t, \mathbf{z}_t$ are sampled from the joint distribution: $\{(\mathbf{x}^{(1)}_t, \mathbf{z}^{(1)}_t)$, $\cdots$ $(\mathbf{x}^{(n)}_t, \mathbf{z}^{(n)}_t)\}$ $\sim$ $\mathbb{P}_{X_TZ_T}$; $\mathbf{\bar{z}}_t$ is sampled from the marginal distribution $\{\mathbf{\bar{z}}^{(1)}_t$, $\cdots$, $\mathbf{\bar{z}}^{(n)}_t\} \sim \mathbb{P}_{Z_T}$.
The detailed implementation of our proposed model is summarized in Algorithm \ref{alg:da-mine}.

\section{Experiments}


\subsection{Experimental Setup}
For the evaluation on the traditional UDA tasks, we use three benchmark datasets: 1) \textbf{Office-Home} contains 15,500 images of everyday objects \citep{venkateswara2017deep}.The images with different traits/background styles are divided into four different domains and each domain has 65 object classes: Art (\textbf{Ar}), Clipart (\textbf{Cl}), Product (\textbf{Pr}) and Real-World (\textbf{Rw}); 2) \textbf{ImageCLEF-DA}\footnote{https://www.imageclef.org/2014/adaptation} is a dataset used for the 2014 ImageCLEF domain adaptation challenge, which
consists of 12 object classes from three public datasets: \emph{Caltech-256} (\textbf{C}), \emph{ImageNet ILSVRC2012} (\textbf{I}) and \emph{Pascal VOC 2012} (\textbf{P}). Each domain contains 600 images and each class has 50 images; 
3) \textbf{Office-31} is a standard benchmark dataset that contains 31 object classes related to the office environment \citep{saenko2010adapting}. This dataset has three domains: \emph{Amazon} (\textbf{A}), \emph{Webcam} (\textbf{W}), and \emph{DSLR} (\textbf{D}). \emph{Amazon} consists of 2817 images from amazon.com. \emph{Webcam} (795 images) and \emph{DSLR} (498 images) contain  images captured by a web camera and a digital SLR camera, respectively. 

All experiments were implemented on \textbf{PyTorch}\footnote{https://pytorch.org/} platform. In all experiments, we employed the Resnet-50  \citep{he2016deep} architecture, which follows the standard evaluation protocols for UDA as \citep{ganin2014unsupervised, cao2018partial, tzeng2017adversarial} to utilize all samples from the two domains. We repeated each transfer task three times to report the average accuracy and the standard error.
We utilized the SGD optimizer and the unified hyper-parameters for all experiments, with $\alpha=1.0$, $\beta=0.01$, learning rate at 1e-3, and batch size at 32. For fair comparisons, our backbone network is identical to the baseline methods, which is fine-tuned from the ImageNet \citep{deng2009imagenet} pre-trained model. In this work, we tuned the hyperparameters through grid search. We essentially tuned the three hyper-parameters: 1) learning rate was tuned from 1e-4 to 0.1; 2) $\alpha$ was tuned from 1e-2 to 100; 3) $\beta$ was tuned from 1e-4 to 1. Besides, we minimized the conditional entropy of the softmax predictions for the target samples in all experiments, which could encourage our model to better utilize these unlabeled images \citep{grandvalet2005semi}. This penalty can be expressed as $\mathcal{L}_{ent} = \frac{1}{|X_T|}\sum_{\mathbf{x}_t\in X_T}-F(G(\mathbf{x}_t))\log F(G(\mathbf{x}_t))$.



\subsection{Result Analysis}
\textbf{Office-Home} The results of the evaluation on Office-Home are presented in Table \ref{tab:officehome}. Our model outperforms the baseline methods significantly in most transfer tasks, and achieves the highest average classification accuracy. Besides, the improvement of performance is more significant in Office-Home compared with that of other datasets. This is because Office-Home has more significant image variations among different domains. By preserving the representation that is private to the target domain, more useful information can be obtained. This improvement also suggests that the importance of preserving the representation that is private to the target domain for an UDA problem and DPN can perform more effectively in these more challenging tasks.

\begin{table*}[h]
\setlength{\tabcolsep}{6pt}
\setlength\aboverulesep{0pt}\setlength\belowrulesep{0pt}
\caption{Accuracy(\%) of DPN on \emph{Office-Home}}
\vspace{-3mm}
\label{tab:officehome}
\small
\begin{center}
\setlength\tabcolsep{0.8pt}
\begin{tabular}{l cccccccccccc|c}
\toprule

\textbf{Method} & \textbf{Ar}$\veryshortarrow$\textbf{Cl} & \textbf{Ar}$\veryshortarrow$\textbf{Pr} & \textbf{Ar}$\veryshortarrow$\textbf{Rw} & \textbf{Cl}$\veryshortarrow$\textbf{Ar} & \textbf{Cl}$\veryshortarrow$\textbf{Pr} & \textbf{Cl}$\veryshortarrow$\textbf{Rw} & \textbf{Pr}$\veryshortarrow$\textbf{Ar} & \textbf{Pr}$\veryshortarrow$\textbf{Cl} & \textbf{Pr}$\veryshortarrow$\textbf{Rw} & \textbf{Rw}$\veryshortarrow$\textbf{Ar} & \textbf{Rw}$\veryshortarrow$\textbf{Cl} & \textbf{Rw}$\veryshortarrow$\textbf{Pr} & \textbf{Avg} \\
\hline
\hline
ResNet50 \citep{he2016deep}& 34.9 & 50.0 & 58.0 & 37.4 & 41.9 & 46.2 & 38.5 & 31.2 & 60.4 & 53.9 & 41.2 & 59.9 & 46.1\\
\hline
DANN \citep{ganin2014unsupervised} & 45.6 & 59.3 & 70.1 & 47.0 & 58.5 & 60.9 & 46.1 & 43.7 & 68.5 & 63.2 & 51.8 & 76.8 & 57.6 \\
JAN \citep{pu2018jointgan} & 45.9 & 61.2 & 68.9 & 50.4 & 59.7 & 61.0 & 45.8 & 43.4 & 70.3 & 63.9 & 52.4 & 76.8 & 58.3\\
EasyTL \citep{wang2019easy}&\textbf{52.8}&	72.1&	75.9&	55.0&	65.9&	67.6&	54.4&	46.9&	74.7&	63.8&	52.3&	78.0&	63.3\\
SAFN \citep{xu2019larger} & 52.0 & 71.7 & 76.3 & 64.2 & 69.9 & 71.9 & 63.7 & 51.4 & 77.1 & 70.9& 57.1 & 81.5 & 67.3\\
\hline
DPN (Ours)& {51.8}&	\textbf{75.3}&	\textbf{79.4}&	\textbf{66.6}&	\textbf{74.8}&	\textbf{74.6}&	\textbf{63.8}&	\textbf{51.7}&	\textbf{81.5}&	\textbf{74.0}&	\textbf{58.0}&	\textbf{84.3}&	\textbf{69.7}\\
& $\pm$0.1 & $\pm$0.2 & $\pm$0.1 & $\pm$0.3 & $\pm$0.1 & $\pm$0.1 & $\pm$0.5 & $\pm$0.1 & $\pm$0.1 & $\pm$0.2 & $\pm$0.3 & $\pm$0.1 & \\
\bottomrule
\end{tabular}
\end{center}
\vspace{-3mm}
\end{table*}

\textbf{Office-31} The results of the evaluation on Office-31 are presented in Table \ref{tab:office31}. Our model achieves the overall best classification accuracy and outperforms the baseline algorithms by a significant margin in most transfer tasks, except for the ones with \emph{Amazon} to be the target domain. This is because of the \textit{label pollution issue} suggested in \citep{bousmalis2016domain}. Overall, the performance in this dataset suggests that DPN can improve the transferability by preserving the representation that is private to the target domain.

\begin{table*}[h]
\caption{Accuracy(\%) of DPN on \emph{Office-31}.}
\vspace{-3mm}
\label{tab:office31}
\small
\begin{center}
\setlength\tabcolsep{1.5pt}
\setlength\aboverulesep{0pt}\setlength\belowrulesep{0pt}
\begin{tabular}{l c c c c c c | c}
\toprule
\textbf{Method}& \textbf{A}$\rightarrow$\textbf{W} & \textbf{D}$\rightarrow$\textbf{W} & \textbf{W}$\rightarrow$\textbf{D} & \textbf{A}$\rightarrow$\textbf{D} & \textbf{D}$\rightarrow$\textbf{A} & \textbf{W}$\rightarrow$\textbf{A} & \textbf{Avg} \\
\midrule
\hline
ResNet-50 \citep{he2016deep} & 68.4 & 96.7 & 99.3 & 68.9& 62.5 &60.7 & 76.1\\
\hline
DANN \citep{ganin2014unsupervised} & 82.0 & 96.9 & 99.1 & 79.7& 68.2 & 67.4& 82.2 \\
JAN \citep{pu2018jointgan} & 85.4 & 97.4 & 99.8 & 84.7& 68.6 & 70.0 & 84.3 \\
GTA \citep{sankaranarayanan2018generate} & 89.5 & 97.9 & 99.8 & 87.7& 72.8 & {71.4} & 86.5 \\
MSTN \citep{xie2018learning} & 80.5$\pm$0.4&	96.9$\pm$0.1&	99.9$\pm$0.1&	74.5$\pm$0.4&	62.5$\pm$0.4&	60.0$\pm$0.6 & 79.1\\
DupGAN \citep{hu2018duplex} &73.2$\pm$0.2&-	&-	&	74.1$\pm$0.6&	61.5$\pm$0.5&	59.1$\pm$0.5&-\\
CCN\citep{hsu2018learning} &78.2&	97.4&	98.6&	73.5&	62.8&	60.6&78.5\\
SAFN \citep{xu2019larger} & 90.1$\pm$0.8 & 98.6$\pm$0.2 & 99.8$\pm$0.0 & 90.7$\pm$0.5& \textbf{73.0}$\pm$0.2 & \textbf{70.2}$\pm$0.3 & 87.1\\
\hline
DPN (Ours) & \textbf{91.5}$\pm$0.4 & \textbf{99.5}$\pm$0.5& \textbf{100.0}$\pm$0.0& \textbf{94.0}$\pm$0.9& 72.2$\pm$1.3 & {68.1}$\pm$0.1 & \textbf{87.6}\\
\bottomrule
\vspace{-5mm}
\end{tabular}
\end{center}
\end{table*}

\textbf{ImageCLEF-DA} The results of the evaluation on ImageCLEF-DA are presented in Table \ref{tab:imageclef}. It is seen that our model outperforms the baseline methods by a large margin in the adaptation scenarios \textbf{I} $\rightarrow$ \textbf{P}, \textbf{P} $\rightarrow$ \textbf{C}, and \textbf{I} $\leftrightarrow$ \textbf{C}, and achieves the best average classification accuracy.

\begin{table*}[h]
\setlength{\tabcolsep}{2pt}
\caption{Accuracy(\%) of DPN on \emph{ImageCLEF-DA}.}
\vspace{-3mm}
\setlength\aboverulesep{0pt}\setlength\belowrulesep{0pt}
\label{tab:imageclef}
\small
\begin{center}
\begin{tabular}{l c c c c c c | c}
\toprule
\textbf{Method}& \textbf{I}$\rightarrow$\textbf{P} & \textbf{P}$\rightarrow$\textbf{I} & \textbf{I}$\rightarrow$\textbf{C} & \textbf{C}$\rightarrow$\textbf{I}& \textbf{C}$\rightarrow$\textbf{P} & \textbf{P}$\rightarrow$\textbf{C}&\textbf{Avg}\\
\hline
\hline
ResNet-50 \citep{he2016deep}& 74.8 & 83.9 & 91.5 & 78.0 & 65.5 & 91.2 & 80.7\\
\hline
DANN \citep{ganin2014unsupervised}& 75.0 & 86.0 & 96.2 & 87.0 & 74.3 & 91.5 & 85.0 \\
JAN \citep{pu2018jointgan}& 76.8 & 88.0 & 94.7 & 89.5 & 74.2 & 91.7 & 85.8 \\
MSNT\citep{xie2018learning}& 67.3$\pm$0.3	&82.8$\pm$0.2	&91.5$\pm$0.1 	&81.7$\pm$0.3	&65.3$\pm$0.2&	91.2$\pm$0.2&80.0\\
EasyTL\citep{wang2019easy}&78.7&	90.3&	96.0&	91.5&	\textbf{77.7}&	95.0&	88.2\\
SAFN \citep{xu2019larger} & 79.3$\pm$0.1 & \textbf{93.3}$\pm$0.4 & 96.3$\pm$0.4 & 91.7$\pm$0.0 & 77.6$\pm$0.1 & 95.3$\pm$0.1 & 88.9\\
\hline
DPN (Ours)&\textbf{80.0} $\pm$ 0.1	&92.9$\pm$0.1	&\textbf{96.8}$\pm$0.1	&\textbf{92.0}$\pm$0.5	&77.0$\pm$0.3	&\textbf{95.7}$\pm$0.2  	&\textbf{89.1}\\
\bottomrule
\vspace{-3mm}
\end{tabular}
\end{center}
\end{table*}

\begin{figure}[h]
\vspace{-2mm}
    \center
    \begin{subfigure}[b]{0.23\textwidth}
        \includegraphics[width=\textwidth]{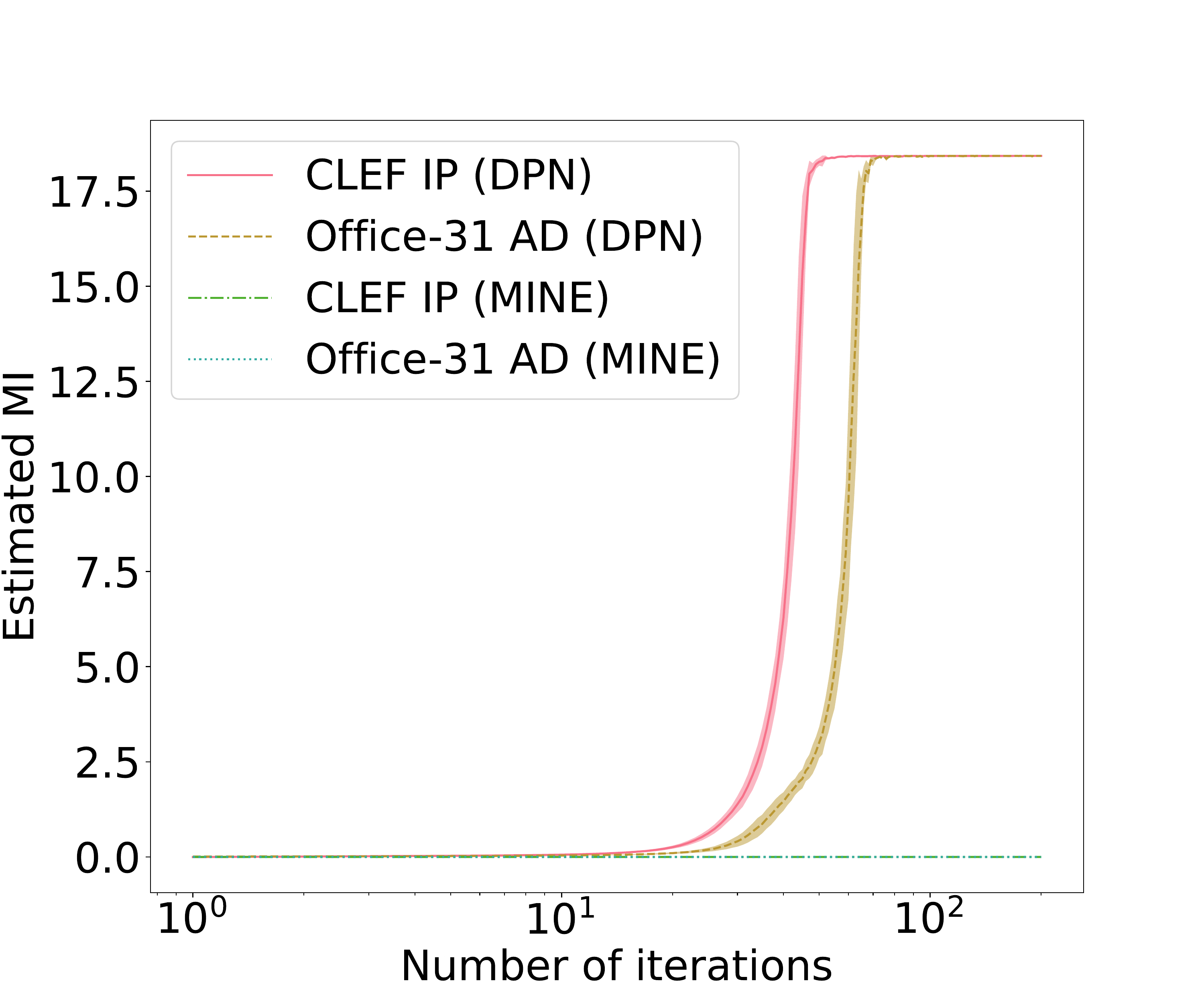}
        \caption{Estimated MI.}
    \end{subfigure}
    \begin{subfigure}[b]{0.23\textwidth}
        \includegraphics[width=\textwidth]{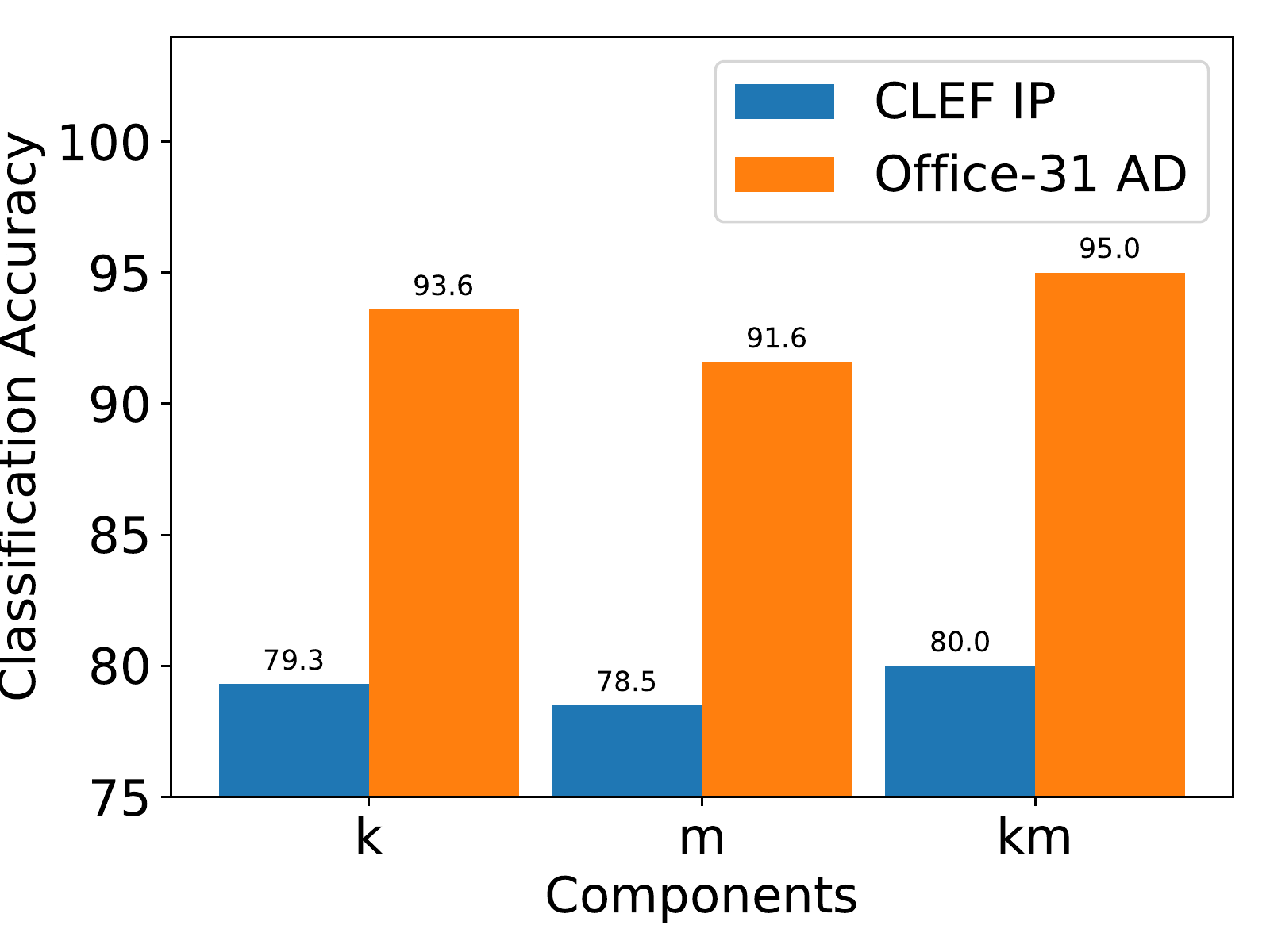}
        \caption{DPN components.}
    \end{subfigure}
    \begin{subfigure}[b]{0.25\textwidth}
        \includegraphics[width=\textwidth]{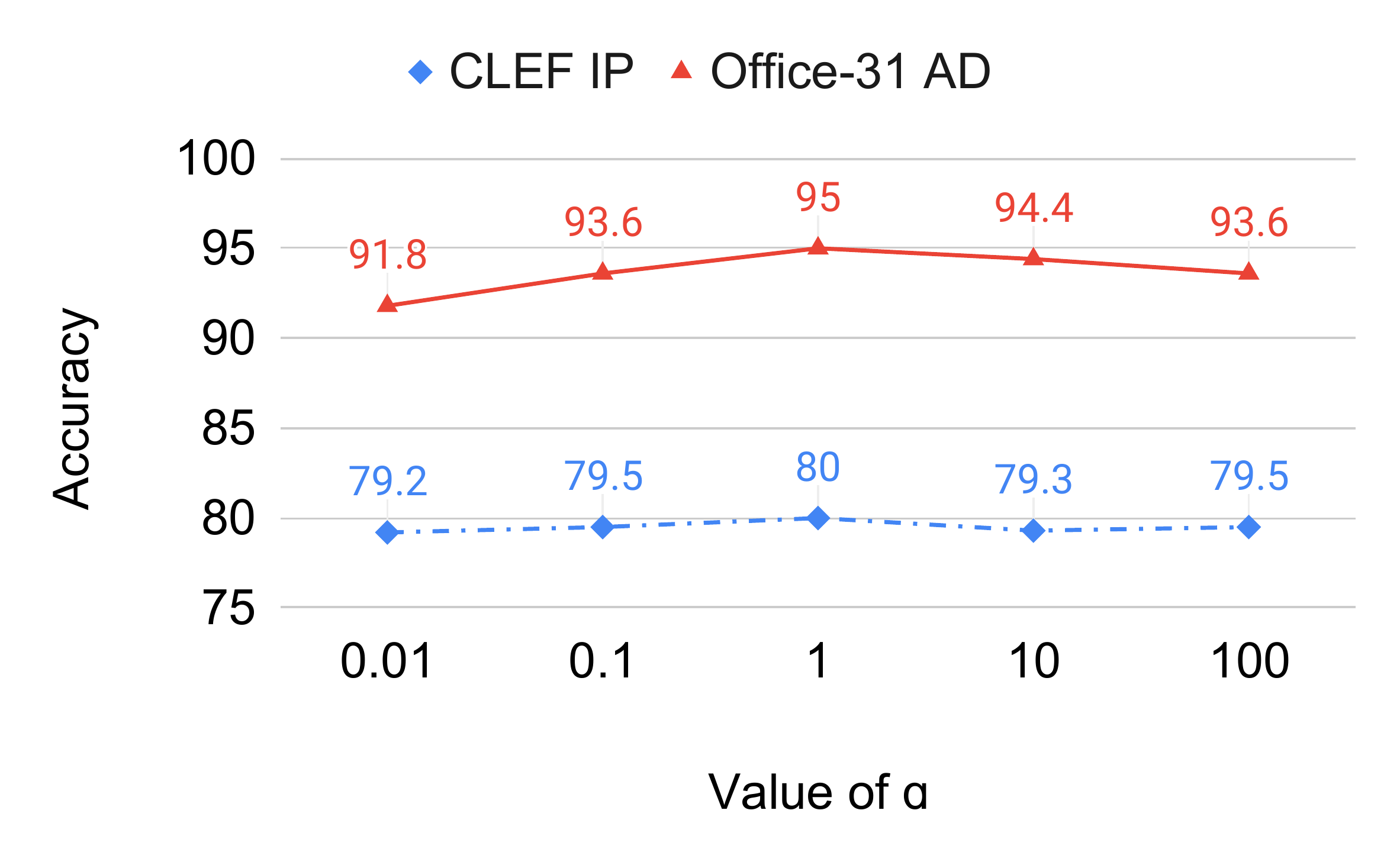}
        \caption{Sensitivity of $\alpha$.}
    \end{subfigure}
    \begin{subfigure}[b]{0.25\textwidth}
        \includegraphics[width=\textwidth]{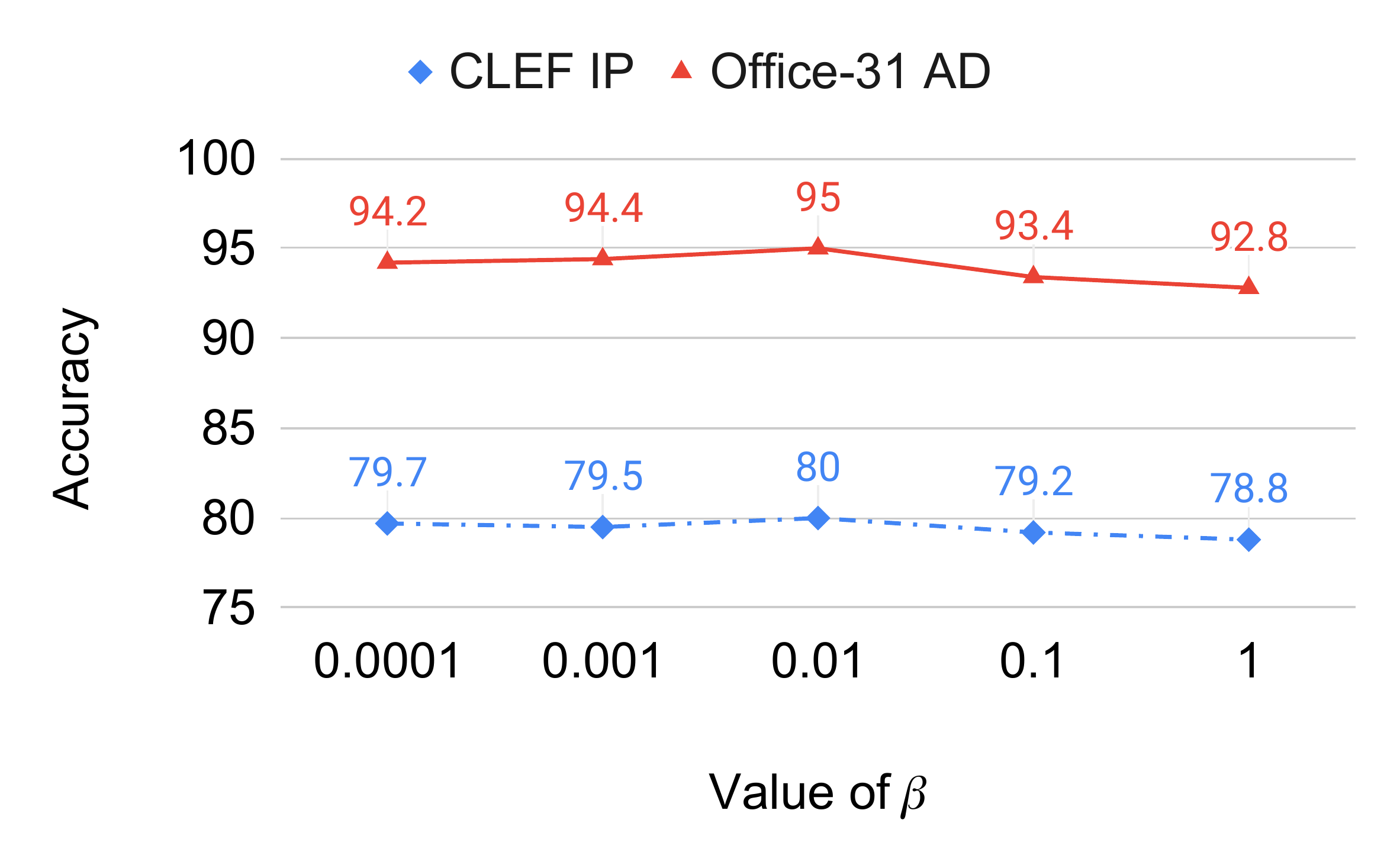}
        \caption{Sensitivity of $\beta$.}
    \end{subfigure}
    \caption{Analysis of (a) convergence of the estimated mutual information; (b) effectiveness of each component of DPN; (c)(d) parameter sensitivities of $\alpha$ and $\beta$.}\label{fig:abl}
\vspace{-3mm}
\end{figure}

\textbf{Convergence of Mutual Information: }Figure \ref{fig:abl} (a) presents the convergence of the estimated mutual information between the target input space and its latent representation. The estimated MI converged to around 18 within 100 epochs in both cases, which indicates that DPN can estimate the MI effectively and efficiently. In comparison, if the joint probability and the product of the marginals are estimated jointly by one network as proposed in MINE \citep{belghazi2018mutual}, the MI is almost impossible to be estimated. As shown Figure \ref{fig:abl} (a), the MI estimated by MINE converges to 0.1, which is not ideal. 
\subsection{Ablation study}
\textbf{Component Analysis}: Figure \ref{fig:abl} (b) presents the contribution of each DPN component to the overall performance. For $x \in \{\textrm{k}, \textrm{m}, \textrm{km}\}$, DPN-$x$ denotes the backbone network with only the component $x$, where k denotes the KL-divergence penalty between $Z_S$ and $Z_T$ and m denotes the MI for preserving the representation that is private to the target domain. DPN-m performs the worst suggesting that preserving the domain private representation alone cannot effectively transfer the knowledge. DPN-k achieves better performance than DPN-m, which indicates the effectiveness of knowledge transfer via minimizing the KL-divergence between the latent distributions of the two domains. The best performance is achieved by using the KL-divergence penalty together with the MI regularization. This suggests the importance of preserving the representation that is private to the target domain during the process of the knowledge transfer.

\textbf{Sensitivity of $\alpha$ and $\beta$}: 
Figure \ref{fig:abl} (c) and Figure \ref{fig:abl} (d) illustrate the sensitivity of $\alpha$ and $\beta$ by varying $\alpha \in \{0.01, 0.1, 1, 10, 100\}$ and $\beta \in\{0.0001, 0.001, 0.01, 0.1, 1\}$. For both parameters, the performance almost stays the same as the parameters vary, which suggests that DPN can work robustly in solving the UDA problem.

\textbf{How to effectively maximize the mutual information}: Table \ref{tab:compare-to-AE} presents the overall performance of maximizing the MI by the autoencoder-based reconstruction (AE), instead of explicitly estimating the MI using deep neural networks. We also conducted experiments on estimating the MI by MINE \citep{belghazi2018mutual}, and listed their results in Table \ref{tab:compare-to-AE}. The results indicate that our proposed method is the most effective way to maximize the MI for preserving the domain private representation.

\begin{table*}[h]
\setlength{\tabcolsep}{1pt}
\caption{Comparison of using different methods to maximize the mutual information. }
\vspace{-3mm}
\setlength\aboverulesep{0pt}\setlength\belowrulesep{0pt}
\label{tab:compare-to-AE}
\small
\begin{center}
\begin{tabular}{l |c c c c c c | c | c c c c c c | c|}
\toprule
\multirow{2}{*}{\textbf{Method}}&\multicolumn{6}{c}{CLEF}&&\multicolumn{6}{c}{Office-31}& \\
\cline{2-15}
& \textbf{I}$\rightarrow$\textbf{P} & \textbf{P}$\rightarrow$\textbf{I} & \textbf{I}$\rightarrow$\textbf{C} & \textbf{C}$\rightarrow$\textbf{I}& \textbf{C}$\rightarrow$\textbf{P} & \textbf{P}$\rightarrow$\textbf{C}&\textbf{Avg}&\textbf{A}$\rightarrow$\textbf{W} & \textbf{D}$\rightarrow$\textbf{W} & \textbf{W}$\rightarrow$\textbf{D} & \textbf{A}$\rightarrow$\textbf{D} & \textbf{D}$\rightarrow$\textbf{A} & \textbf{W}$\rightarrow$\textbf{A} & \textbf{Avg}\\
\hline
KLD + AE & 79.2 &	92.5&	96.3&	91.5&	76.0&	94.7& 88.4&90.6 & 99.1& 100.0& 93.0 & 70.9 &  67.4& 86.8\\
KLD + MINE & 79.5 &	92.3&	96.3&	91.3&	76.3&	94.8& 88.4&90.7 & 99.0& 100.0& 92.2 &  69.9& 66.8 & 86.4\\
DPN (Ours)&\textbf{80.0} & \textbf{92.9} & \textbf{96.8} &\textbf{92.0} & \textbf{77.0} & \textbf{95.7} & \textbf{89.1}& \textbf{91.1} & \textbf{99.2}& 100.0& \textbf{95.0} & \textbf{73.4}&\textbf{68.1}&\textbf{87.8}\\
\bottomrule
\end{tabular}
\end{center}
\end{table*}

\subsection{Adaptability to the Third Dataset}
\begin{figure}[h]
\vspace{-4mm}
    \center
    \captionsetup[subfigure]{width=1.5\linewidth}
    \begin{subfigure}[b]{0.42\textwidth}
        \includegraphics[width=\textwidth]{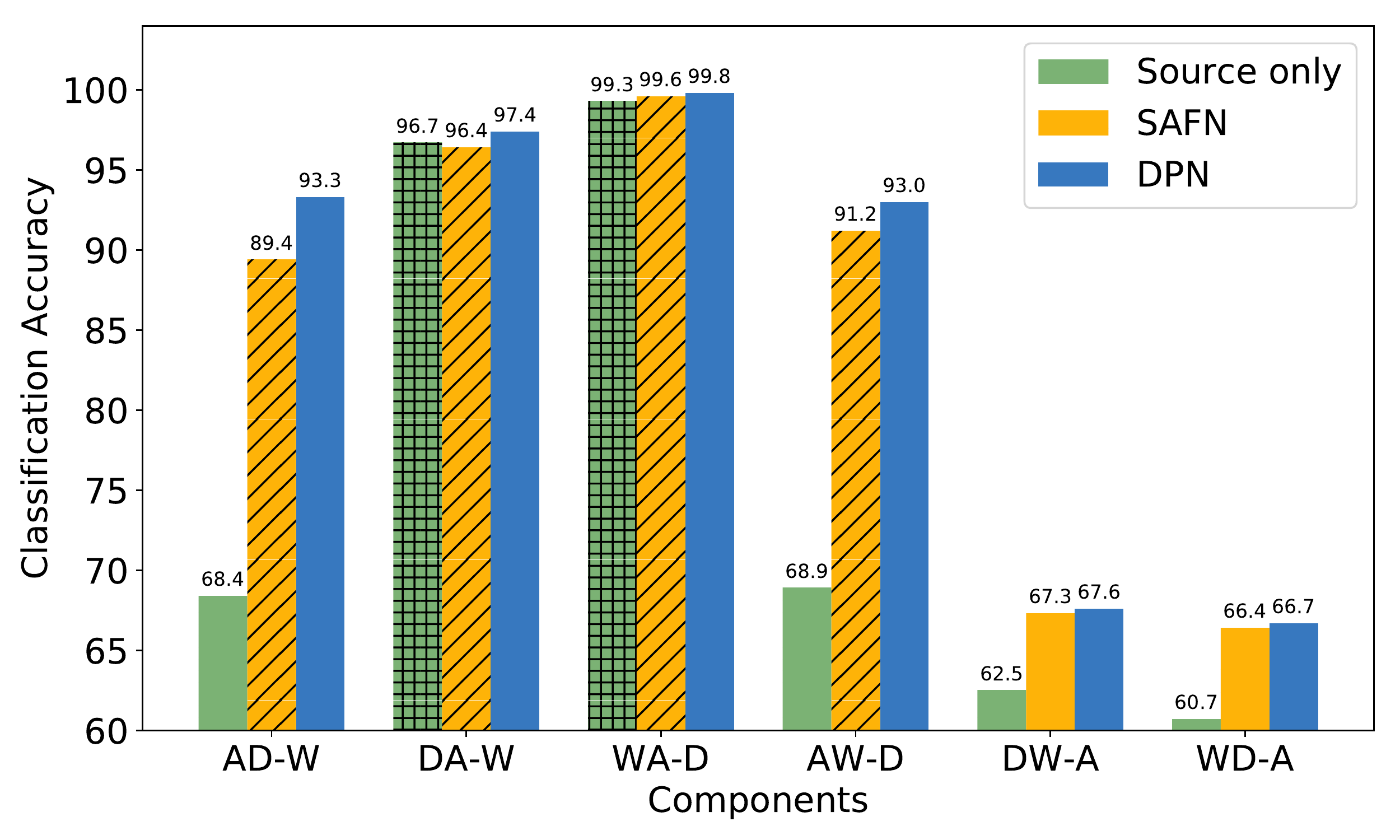}
        \caption{Generalization comparison.}
    \end{subfigure}\hfill
    ~~~~
    \begin{subfigure}[b]{0.42\textwidth}
        \includegraphics[width=\textwidth]{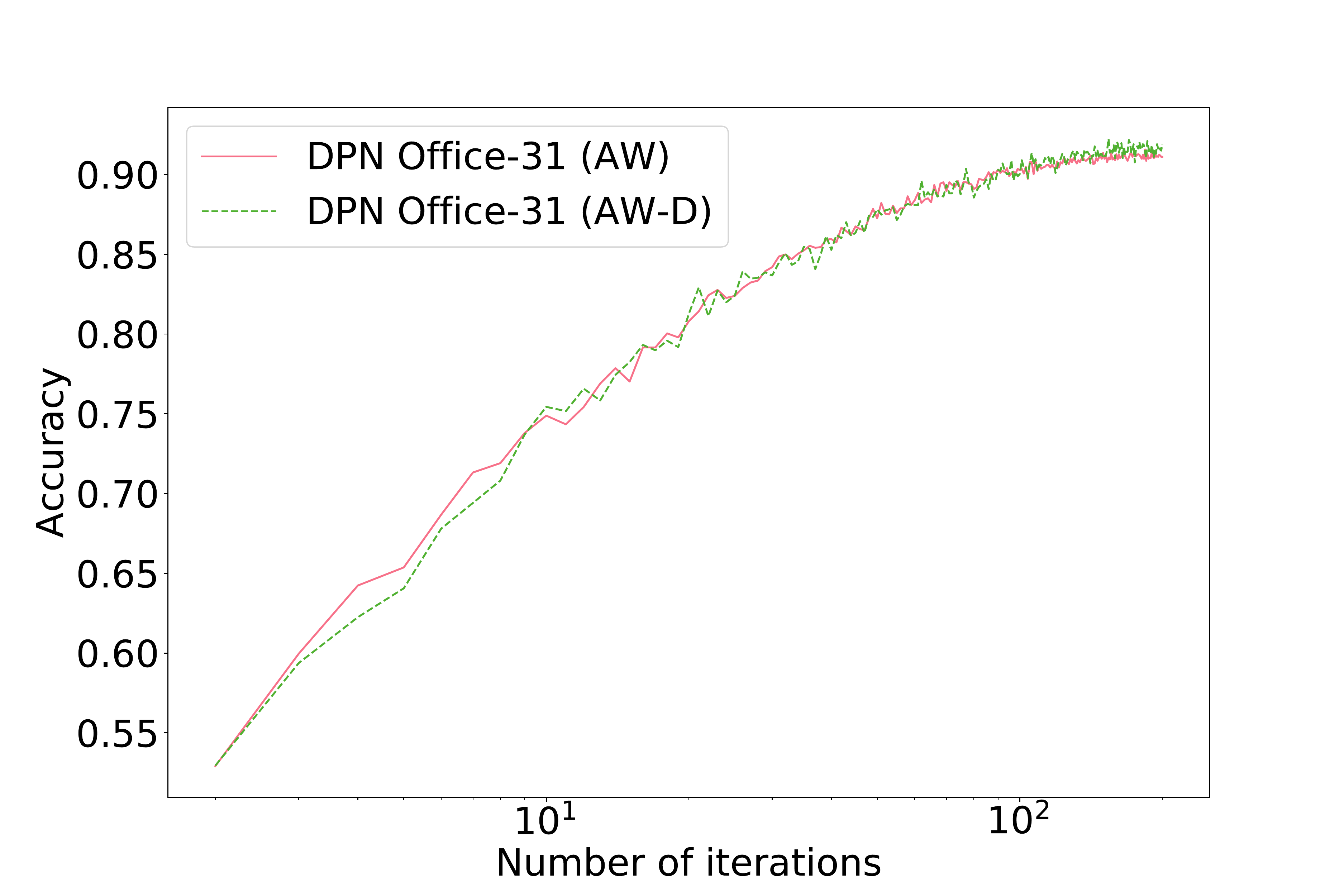}
        \caption{Relationship between generalization and adaptation.}
    \end{subfigure}
    \caption{Adaptability to the Third Dataset on Office-31.}\label{fig:thrid}
\end{figure}

To better validate the generalization and adaptability of our model to an unknown scenario, we utilized the best-performing models, which were trained previously, and tested their performance on a new dataset. For example, we tested the model trained on the adaptation scenario from \textbf{A} to \textbf{D} on a third dataset \textbf{W}, and denoted the transfer task as \textbf{AD-W}. Note that the trained models do not have any prior information, i.e., neither the data samples nor the labels, about this third dataset. We conducted a case study on Office-31 \citep{saenko2010adapting}, and compared the generalization of our model with that of SAFN \citep{xu2019larger}. Figure \ref{fig:thrid} (a) shows the comparison of the generalization to the third dataset. It is seen that the domain adaptation strategies can generally improve the generalization of a discriminative model, except for the poor adaptability of SAFN to \textbf{DA-W}. Moreover, the proposed DPN outperforms SAFN in all cases, which indicates that preserving the representation that is private to the target domain can significantly improve the generalization of a discriminative model. Figure \ref{fig:thrid} (b) presents how the generalization to the third dataset improves with the progress of adaptation. We demonstrated this on \textbf{AW-D}. The performance on the third dataset, \emph{DSLR}, has strong correlation with that on the \emph{Webcam}. The Pearson's correlation coefficient between two accuracy curves in Figure \ref{fig:thrid} (b) is 0.99, which means they have almost total positive linear correlation.

\section{Conclusion}
In this paper, we proposed a novel method to preserve the domain private representation to improve the generalization of the UDA models. We demonstrated both theoretically and empirically that preserving such representation is important for the knowledge transfer. We also showed that the joint probability and the product of the marginals were better to be estimated by two separate deep neural nets in the setting of UDA. Experimental results have shown the significance of the proposed research by achieving state-of-the-art performance in several benchmark datasets.

\section*{Broader Impact}

The proposed work improves the performance of UDA by preserving the representation that is private to the unlabeled domain, which makes the decision maker more generalize on a new dataset. Recent research in computer vision leads to the application of facial recognition and criminal identification, which arises the ethical concerns on right to privacy and false alters. The adaptability of discriminative models is important when utilizing such model to replace human decisions because the poor generalization to real-world scenarios will result in high false position rates in detecting crimes, which troubles innocent people. Besides, the work involving the real-world personal data is a potential for the sensitive-personal-information leak. Therefore, the research on improving the adaptability of discriminative models by 
making the most of the limited data is of necessity. 

\bibliography{bibtex} 
\end{document}